%% file: main.tex
\documentclass{article}
\usepackage{iclr2027_conference,times}
\usepackage{amsmath,amssymb,amsthm,booktabs,graphicx,xcolor}
\usepackage{tabularx,placeins}
\usepackage{url}
\usepackage{multirow} 
\usepackage{float}
\usepackage{hyperref}
\usepackage{sopsd_preprint}
\usepackage{fontawesome5}
\hypersetup{colorlinks=true,linkcolor=ReferenceBlue,citecolor=ReferenceBlue,urlcolor=ReferenceBlue,
  pdftitle={Shuffled OPSD: When Semantic Consistency Hurts On-Policy Self-Distillation?},
  pdfauthor={Qingyuan Zheng and Zhihao Dou}}
\title{Shuffled OPSD: When Semantic Consistency Hurts On-Policy Self-Distillation?}
\author{Qingyuan Zheng, Zhihao Dou\\Duke University\\\texttt{\{pearson.z,zhihao.dou\}@duke.edu}}
\iclrfinalcopy
\newcommand{\E}{\mathbb E}

\begin{document}
\maketitle
\sopsdpreprintheader
\vspace{-20pt}
\begin{center}
\href{https://github.com/Pear3on/S_OPSD}{{\color{black}\faGithub}\enspace GitHub}
\end{center}
\vspace{0pt}
\input{sections/abstract}
\input{sections/teaser}
\clearpage

\input{sections/introduction}

\input{sections/preliminaries}

\input{sections/design}

\input{sections/results}

\input{sections/analysis}
\input{sections/related_work}

\input{sections/conclusion}

\clearpage

\input{sections/ai_use}

\bibliography{references}
\bibliographystyle{iclr2027_conference}
\clearpage
\appendix
\input{sections/appendix}
\end{document}

%% file: sections/abstract.tex
\begin{abstract}
Knowledge distillation improves a student model's reasoning ability through supervision from a teacher. On-policy self-distillation (OPSD) uses the same model as both teacher and student under different contexts: the student generates reasoning trajectories from the question alone, while the teacher additionally receives a complete worked example and provides token-level supervision along those trajectories. Because the problem statement is shared, its content and ordering shape both student generation and teacher feedback. Preserving its semantic structure therefore seems natural, but does it always improve learning? We introduce \emph{Shuffled OPSD}, which shuffles the words of the shared problem statement while keeping the teacher-only worked example intact. Across three mathematical reasoning benchmarks and three Qwen3 model scales, the method improves mean accuracy over both an intact-target control and Standard OPSD under matched training configurations. The largest gain over Standard OPSD is 3.52 percentage points on Qwen3-1.7B. Further controls show that preserving larger word blocks offers no advantage over word-level shuffling at the evaluated scales, whereas shuffling the teacher-only worked example reduces accuracy at every scale. Cross-problem mixing and target-removal experiments further distinguish the roles of problem content, word order, and teacher context. These results show that disrupting the shared problem statement can improve learning, while disrupting the teacher-only reference hurts performance. They challenge semantic coherence as a universally beneficial principle for OPSD and motivate closer study of how input structure shapes student generation and teacher supervision.
\end{abstract}

%% file: sections/teaser.tex
\begin{figure}[H]
\centering
\includegraphics[width=\linewidth]{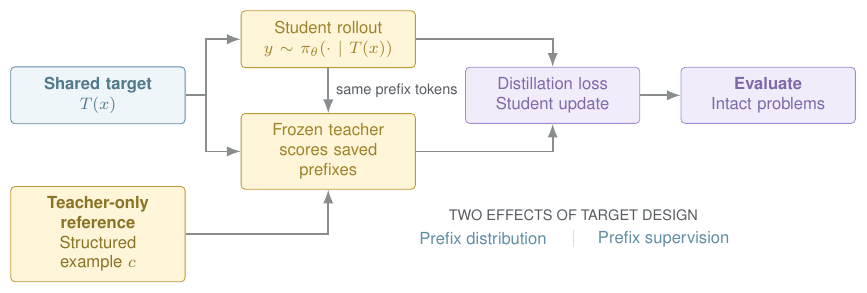}
\caption{
Shuffled OPSD transforms the problem statement shared by student and
teacher during training.
The transformation affects both student rollout generation and
frozen-teacher scoring of the same prefixes, while the teacher-only
reference remains intact.
Evaluation uses the final step-50 student on the original unmodified
problems.
}
\label{fig:pipeline}
\end{figure}

%% file: sections/introduction.tex
\section{Introduction}

Recent advances in large language models (LLMs) have substantially improved mathematical reasoning through post-training methods such as reinforcement learning \citep{shao2024deepseekmath,yu2026dapo,liu2025understanding}, supervised fine-tuning \citep{ouyang2022training,wei2021finetuned}, and knowledge distillation \citep{R01}.
Among these approaches, on-policy distillation combines two useful properties: the student learns on trajectories sampled from its own policy, while a teacher provides dense token-level supervision on those trajectories \citep{R17,R29}.
On-policy self-distillation (OPSD) \citep{R01} further removes the need for a separate teacher model by instantiating teacher and student from the same backbone under different conditioning contexts.
The student sees the problem and generates its own response, while a frozen teacher scores the same prefixes with access to privileged context such as a worked solution.

Most work on OPSD has focused on the privileged information available to the teacher.
Recent studies examine whether the teacher needs a correct solution, how much reasoning should be exposed, and whether useful learning persists with absent or unrelated references \citep{R03,R04,R06,R10,R11}.
These results suggest that richer or more semantically relevant teacher context does not always translate into a better trained student.
However, a separate part of the OPSD conditioning structure has received much less attention: the problem statement shared by both teacher and student is normally kept intact.

The shared problem plays a different role from teacher-only privileged information.
It determines which trajectories the student generates and therefore which prefixes receive supervision.
It also conditions the student and teacher distributions used to compute the distillation loss.
Its structure thus affects both where learning occurs and what feedback the teacher provides.
This makes the shared problem a distinct design dimension: changing its structure can alter the learning process even when the teacher reference and distillation objective remain fixed.

We ask whether preserving the semantic structure of the shared problem improves OPSD, or whether deliberately disrupting it can produce a better student.
Destroying word order appears counterproductive for mathematical reasoning.
Yet OPSD trains through teacher feedback on student-generated trajectories, rather than directly rewarding the semantic coherence of its inputs.
A problem statement that is easier to interpret need not induce trajectories and supervision that lead to better learning.
This motivates treating the shared problem as a \emph{training interface}, whose structure can differ from that of the intact problems used at inference.

Based on this intuition, we introduce \textbf{Shuffled OPSD}.
We first replace the teacher's target-associated solution with a complete worked example from an unrelated problem, forming the \textsc{UnrelatedRef} condition.
Across three model scales, this condition retains performance comparable to Standard OPSD, with differences in three-benchmark mean accuracy of at most 0.55 percentage points.
Building on this condition, Shuffled OPSD permutes the word order of the problem seen by both student and teacher while keeping the teacher-only reference intact.
The student generates its trajectory from the shuffled problem, and the frozen teacher provides token-level supervision under the same transformed problem together with the reference.
The transformation preserves the problem's words, numbers, and symbolic content while disrupting their ordering.
The distillation objective remains unchanged, and evaluation always uses the original intact problems.

Our experiments reveal a consistent pattern: replacing the teacher reference with an unrelated example preserves performance, shuffling the shared problem improves it, and additionally shuffling the teacher reference reduces it.
Using the final checkpoints after 50 optimizer steps, word-level shuffling improves three-benchmark mean accuracy over Standard OPSD by 3.52, 1.11, and 1.38 percentage points on Qwen3-1.7B, Qwen3-4B, and Qwen3-8B, respectively.
At 1.7B and 4B, preserving larger word blocks offers no advantage over word-level shuffling.
These results support \emph{selective semantic disruption}: disrupting the shared problem's ordering while retaining the structure of the teacher's reference.
They extend the study of OPSD from what information the teacher receives to how the shared input is organized.

\textbf{Our contributions are as follows:}
First, we identify the semantic structure of the shared problem as a distinct design dimension in OPSD, affecting both student-generated trajectories and teacher supervision.
Second, we introduce \textbf{Shuffled OPSD}, a simple input transformation that improves reasoning performance without changing the distillation objective or inference procedure.
Third, through controlled interventions on the shared problem and teacher-only reference, we show that their semantic structures play different roles and establish selective semantic disruption as an effective input-design principle across three model scales.

%% file: sections/preliminaries.tex
\section{Preliminaries}
\label{sec:prelim}
\subsection{On-Policy Self-Distillation}
\label{sec:opsd_mechanics}
In on-policy self-distillation \citep{R17,R01}, a student $\pi_\theta$ generates continuation $y$ for target $x$ in mode $m_S$. A frozen teacher with initial parameters $\theta_0$ scores the same prefixes with reference $c$ and mode $m_T$:
\begin{equation}
\begin{aligned}
p^S_t&=\pi_\theta(\cdot\mid x,y_{<t};m_S),\\
p^T_{c,t}&=\pi_{\theta_0}(\cdot\mid x,c,y_{<t};m_T).
\end{aligned}
\label{eq:distributions}
\end{equation}
Both models see the target, hence \emph{shared target}. The reference provides teacher-only context, rather than a separate output sequence for the student to imitate.

For supervised positions $I(y)$, the training objective is
\begin{equation}
\mathcal L_{I,c}(\theta)=\E_x\E_{y\sim\pi_\theta(\cdot\mid x;m_S)}
\left[\frac{1}{|I(y)|}\sum_{t\in I(y)}
\ell_{\rm cap}(p^T_{c,t},p^S_t)\right].
\label{eq:base_loss}
\end{equation}
The loss $\ell_{\rm cap}$ caps each signed teacher-to-student KL contribution from above at $0.05$ before summation (Appendix~\ref{app:implementation}). It is therefore not ordinary KL. Teacher outputs, sampled tokens, and sampling probabilities are held fixed during differentiation.

\subsection{Shared and Teacher-Only Conditioning}
\label{sec:conditioning}
The shared target directly conditions generation and scoring. The reference directly conditions only scoring, although the resulting updates can change later student rollouts. Our default student uses non-thinking mode and the teacher uses thinking mode, as specified by their chat templates.

Standard OPSD pairs each target with its associated solution. UnrelatedRef instead uses a worked example from another problem. Fixing this reference assignment allows us to vary the shared target independently. All evaluation uses intact problems without teacher-only context.

\begin{figure}[t]
\centering
\includegraphics[width=\linewidth]{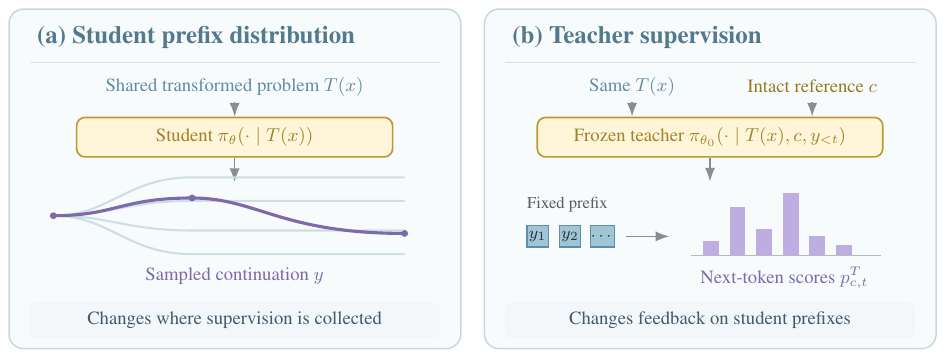}
\caption{
Shuffled OPSD selectively transforms the problem statement shared by
student and teacher.
The transformation changes both the student prefix distribution and the
distillation signal evaluated on those prefixes, while the teacher-only
reference remains intact. Paths and token scores are schematic.
}
\label{fig:selective_disruption}
\end{figure}

%% file: sections/design.tex
\section{Design of Shuffled OPSD}
\label{sec:setup}

\subsection{Semantic Structure as a Design Variable}
\label{sec:design}

OPSD provides the student and teacher with the same problem, while
giving the teacher an additional reference.
We distinguish two aspects of this input structure:
whether the reference answers the target problem, and how the
content within each input is organized.
We first replace the reference with a complete worked example
from another problem.
We then shuffle the shared problem while keeping that example intact.

Let $x$ denote the target problem, $c$ the reference available only
to the teacher, and $T$ a transformation of the shared problem.
The student generates a trajectory
$y \sim \pi_\theta(\cdot \mid T(x);m_S)$.
At each selected position $t$, the student and frozen teacher
distributions are
\begin{align}
p^S_{T,t}
&=
\pi_\theta(\cdot \mid T(x),y_{<t};m_S),\\
p^T_{T,c,t}
&=
\pi_{\theta_0}(\cdot \mid T(x),c,y_{<t};m_T).
\end{align}
Training minimizes
\begin{equation}
\mathcal{L}_{T,c}(\theta)
=
\mathbb{E}_{x}
\mathbb{E}_{y\sim\pi_\theta(\cdot\mid T(x);m_S)}
\left[
\frac{1}{|I(y)|}
\sum_{t\in I(y)}
\ell_{\mathrm{cap}}
\left(
p^T_{T,c,t},
p^S_{T,t}
\right)
\right].
\label{eq:opsd_loss}
\end{equation}

The reference $c$ changes the feedback provided by the teacher.
The transformation $T$ changes both the student's rollout
distribution and the predictions of both models on the resulting
prefixes.
We therefore treat reference selection and shared problem structure
as separate design choices.

\begin{figure}[t]
    \centering
    \includegraphics[width=0.94\linewidth]
    {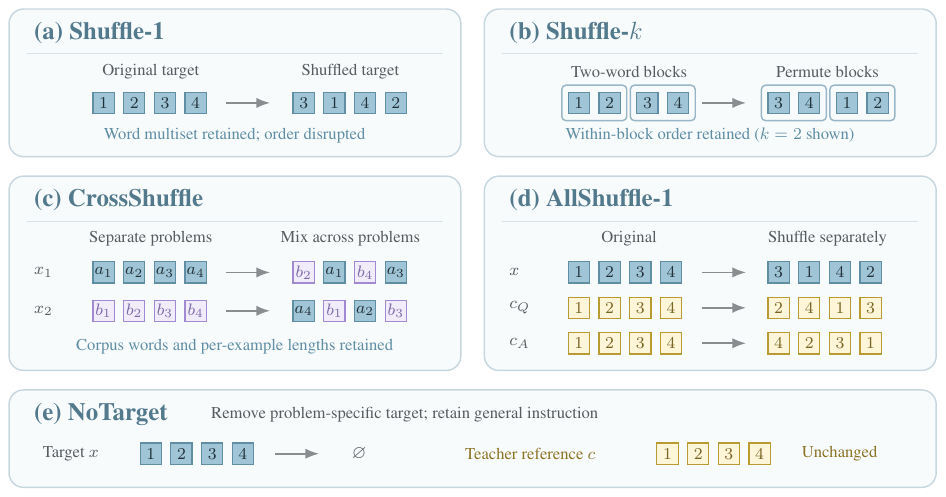}
    \caption{
    \textbf{Input interventions in Shuffled OPSD.}
    UnrelatedRef replaces the original teacher reference with an
    intact worked example from another problem.
    Shuffle-$k$ additionally permutes blocks within the shared
    problem.
    CrossShuffle mixes words across problems, and NoTarget removes
    the specific problem statement.
    AllShuffle additionally shuffles the reference provided to
    the teacher.
    Here $c_Q$ and $c_A$ denote the reference question and solution.
    Permutations are illustrative.
    }
    \label{fig:shuffle_operations}
\end{figure}

\subsection{UnrelatedRef}
\label{sec:unrelatedref}

Standard OPSD uses the intact problem $T_I(x)=x$ together with
a reference solution associated with that problem,
$c_{\mathrm{assoc}}(x)$.
Our first intervention, \emph{UnrelatedRef}, retains the intact
problem but replaces the reference with a complete worked example
from another problem:
\begin{equation}
c_U(x)
=
\left(
x^{\mathrm{ref}},a^{\mathrm{ref}}
\right),
\qquad
x^{\mathrm{ref}} \neq x.
\end{equation}
Here $x^{\mathrm{ref}}$ and $a^{\mathrm{ref}}$ denote the reference
question and its worked solution.
Both remain intact and are visible only to the teacher.

The reference pairing is fixed during training and reused across
all subsequent interventions on the shared problem.
UnrelatedRef tests whether the teacher needs a solution to the
specific target problem.
It also provides the starting point for Shuffled OPSD.

\subsection{Shuffled OPSD}
\label{sec:method}

Shuffled OPSD builds on UnrelatedRef by shuffling the problem seen
by both student and teacher.
The unrelated worked example provided to the teacher remains intact.
This allows us to change the shared problem without changing the
reference.

We split $x$ at whitespace boundaries and group the resulting words
into consecutive blocks $b_1,\ldots,b_M$, each containing $k$ words.
The final block may contain fewer than $k$ words.
We sample a permutation $\sigma$ and hold it fixed for that example
during training:
\begin{equation}
T_k(x)
=
b_{\sigma(1)}
\Vert
b_{\sigma(2)}
\Vert \cdots \Vert
b_{\sigma(M)},
\label{eq:shuffle}
\end{equation}
where $\Vert$ joins blocks with word separators.

Our main variant, \emph{Shuffle-1}, permutes individual words.
It preserves each word and its frequency, including numerical and
symbolic strings within words, while disrupting their order.
Larger blocks preserve more local order and allow us to test the
value of that structure.

The student generates from $T_k(x)$.
The frozen teacher scores the same continuation using $T_k(x)$
and $c_U(x)$ as context.
The student is updated with
$\mathcal{L}_{T_k,c_U}$ in Equation~\ref{eq:opsd_loss}.
The loss, optimizer, and reference pairing remain unchanged from
UnrelatedRef.
At inference, the student receives the original intact problem
without an additional reference.

This design implements \emph{selective semantic disruption}:
we remove the association between the problem and reference,
disrupt the order of the shared problem, and preserve the internal
structure of the teacher's worked example.

\subsection{Controls on Semantic Structure}
\label{sec:interventions}

We use matched controls to examine reference relevance, problem
content and order, and the internal structure of the reference.

\paragraph{Reference relevance.}
Standard OPSD and UnrelatedRef both retain the intact target problem.
Their comparison measures the effect of replacing the corresponding
solution with an unrelated worked example.
UnrelatedRef and Shuffle-1 use identical reference pairings, so
their comparison measures the effect of shuffling the shared problem.

\paragraph{Problem content and order.}
Shuffle-$k$ varies how much local word order is preserved.
\emph{CrossShuffle} pools words across training problems, globally
permutes the pool, and redistributes words according to the original
word count of each example.
It disrupts both word order and problem membership while preserving
word counts across the corpus.
\emph{NoTarget} removes the specific problem statement and its wrapper
while retaining the general instruction.
All these conditions use the same unrelated reference pairings.

\paragraph{Reference structure.}
\emph{AllShuffle-$k$} applies block shuffling to both the shared
problem and the reference provided to the teacher.
AllShuffle-1 separately permutes words in the target problem,
the reference question, and the reference solution.
Comparing Shuffle-1 with AllShuffle-1 measures the additional effect
of disrupting the reference structure.

These controls distinguish which input structures support learning.
Additional experiments on reference removal and teacher mode appear
in Section~\ref{sec:analysis}.
Implementation details, fixed reference pairings, and permutation
seeds are provided in Appendix~\ref{app:implementation}.

%% file: sections/results.tex
\section{Experiments}
\label{sec:results}

We evaluate Shuffled OPSD on three model scales and three
competition-level mathematical reasoning benchmarks.
Our main experiments compare shared-problem disruption with an
intact-problem control and Standard OPSD.
We then analyze how the effect changes with the amount and location of
semantic structure preserved during training.

\subsection{Experimental Setup}
\label{sec:training_eval}

\paragraph{Models and training.}
We experiment with Qwen3-1.7B, Qwen3-4B, and Qwen3-8B \citep{qwen3}
using LoRA adapters \citep{lora} ($r=64$, $\alpha=128$).
Training examples are drawn from the OpenThoughts mathematical-reasoning
subset used by the original OPSD study.
All conditions use 50 optimizer steps, batch size 16, learning rate
$5\times10^{-6}$, and training seed 42.
The teacher is the frozen initial backbone in thinking mode, while the
student generates in non-thinking mode with rollouts capped at 1,024
tokens.
We report the step-50 checkpoint for all trained conditions.
Additional implementation details appear in
Appendix~\ref{app:implementation}.

\paragraph{Evaluation.}
We evaluate on AIME 2024, AIME 2025, and HMMT February 2025, using
30 problems per benchmark.
Evaluation uses the original intact problems without teacher-only
context.
We sample 12 responses per problem in thinking mode with temperature
$1$, top-$p=0.95$, and a maximum of 38,912 new tokens.
Avg\@12 is the primary metric, and the three-benchmark mean weights all
benchmarks equally.
Compared conditions use the same problem order and generation seed
schedule.

\paragraph{Statistics.}
Paired 95\% confidence intervals are estimated with 10,000
problem-cluster bootstrap resamples, keeping all responses from each
problem in the same cluster and using matched problem indices across
conditions.

\subsection{Main Results}
\label{sec:main_results}

Table~\ref{tab:main_results} shows that Shuffle-1 improves mean
accuracy at all three model scales.
We organize the comparisons around two questions: whether shuffling
the shared problem improves learning, and whether the reference
provided to the teacher should also be shuffled.

\paragraph{Shared problem order is a useful training variable.}
Shuffle-1 and UnrelatedRef use the same reference and differ only
in whether the shared problem is shuffled.
Shuffle-1 improves the three-benchmark mean by 3.98, 0.56, and
1.11 percentage points on Qwen3-1.7B, Qwen3-4B, and Qwen3-8B,
respectively.
At 1.7B and 8B, it improves all three benchmarks.
Because evaluation uses intact problems, these gains show that
disrupting the training input can improve reasoning on normally
structured questions.
The shared problem therefore serves as more than a statement to
be solved: its organization is a training variable that affects
student generation and teacher feedback.

\paragraph{The complete design outperforms Standard OPSD.}
Shuffle-1 exceeds Standard OPSD by 3.52, 1.11, and 1.38 percentage
points across the three scales.
Standard OPSD supplies a reference associated with the target,
whereas Shuffle-1 combines an unrelated reference with a shuffled
problem.
The stronger results show that accurate pairing and intact word
order are not prerequisites for effective distillation.
CrossShuffle also exceeds Standard OPSD at every scale, but remains
below Shuffle-1.
This comparison favors retaining the words of each problem while
changing their order, rather than replacing their original grouping
with words drawn from other problems.
The largest improvement occurs at 1.7B, showing that this input
transformation is particularly effective for the smallest model
in our evaluation.

\paragraph{Reference structure remains valuable.}
AllShuffle-1 additionally shuffles the reference while retaining
the same shuffled problem.
Mean accuracy then falls by 2.41, 0.65, and 1.57 percentage points
at 1.7B, 4B, and 8B.
Thus, the benefit comes from where disruption is applied, rather
than from increasing disruption throughout the input.
The shared problem shapes both the trajectories generated by the
student and the feedback provided by the teacher.
The reference supplies context only to the teacher.
Their different roles are reflected in the results: disrupting
shared problem order improves learning, while preserving reference
structure supports stronger supervision.
This pattern motivates selective semantic disruption as the
central design principle of Shuffled OPSD.

\input{tables/main_res}

\subsection{Analysis and Discussion}
\label{sec:analysis}

The experiments reveal a structural asymmetry in OPSD.
Semantic structure in the shared problem is not uniformly beneficial,
whereas structure in the teacher-only reference remains valuable.
The shared problem determines where supervision is collected; the
reference shapes the teacher signal at those locations.

\subsubsection{How Much Structure Should Be Preserved?}
\label{sec:structure_amount}

Among the tested target-side perturbations, the strongest disruption
performs best.
Shuffle-1 reaches 42.78\% at 1.7B, compared with 41.02\% for
Shuffle-4 and 40.65\% for Shuffle-8, and 63.06\% at 4B, compared with
62.22\% and 62.87\%, respectively.
Thus, preserving more local word order does not improve performance.

CrossShuffle and NoTarget further isolate what should be preserved.
They reach 41.11\%, 62.78\%, and 63.43\%, and
41.76\%, 61.95\%, and 62.22\% across the three scales, respectively,
both below Shuffle-1 at every scale.
This suggests that problem-specific lexical and symbolic content matters
more than intact syntax or coherent surface form.
Full results appear in Appendix~\ref{app:sensitivity}.

\subsubsection{Where Does Disruption Help?}
\label{sec:selective_disruption}

AllShuffle-1 additionally shuffles the teacher-only reference and
reduces mean accuracy by 2.41, 0.65, and 1.57 percentage points at
1.7B, 4B, and 8B.
Thus, the same disruption that helps on the shared problem hurts on the
teacher-only context.
This motivates \emph{selective semantic disruption}: perturb the shared
input that shapes student trajectories while preserving structured
teacher guidance.

At 1.7B, Shuffle-1 raises Avg\@12 from 38.80\% to 42.78\% and
Vote\@12 from 48.89\% to 52.22\%, while Pass\@12 remains at 61.11\%.
At 4B, Avg\@12 rises from 62.50\% to 63.06\% and Vote\@12 from
67.78\% to 72.22\%, while Pass\@12 changes little.
The gain therefore comes mainly from making successful reasoning more
reliable rather than from solving substantially more problems.

\input{tables/controls}

\subsubsection{What Supervision Survives Disruption?}
\label{sec:surviving_supervision}

Under NoTarget, removing the unrelated reference lowers 1.7B accuracy
from 41.76\% to 37.78\%, while switching the teacher to non-thinking
mode lowers it to 38.98\%
(Table~\ref{tab:controls}).
Both reference context and teacher reasoning mode therefore remain
useful even without a coherent shared problem.

We further replay saved Shuffle-1 and CrossShuffle trajectories from
32 questions using the frozen Qwen3-4B model, fixing continuation
tokens while varying only the teacher's reference and thinking mode.

\begin{figure}[t]
\centering
\includegraphics[width=0.85\linewidth]{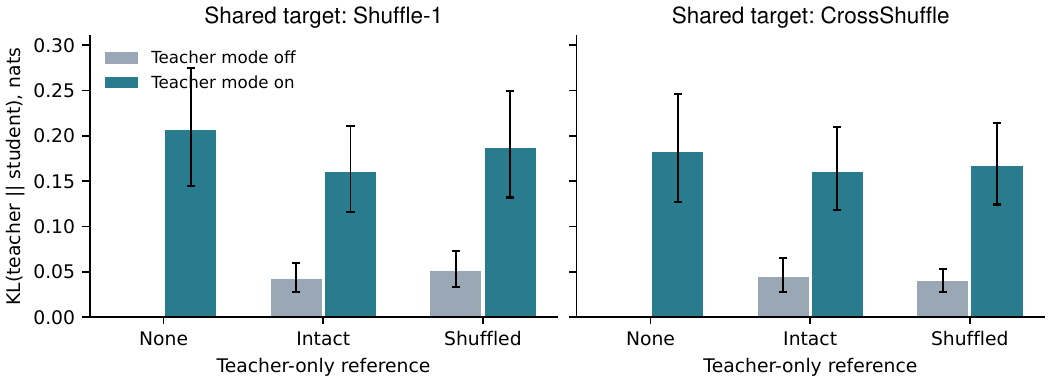}
\caption{
Teacher--student KL on identical prefixes, excluding the first
generated token. Error bars show 95\% bootstrap intervals across
32 questions.
}
\label{fig:mode}
\end{figure}

\input{tables/distribution_core}

Under Shuffle-1, teacher thinking increases KL from 0.0419 to 0.1605
nats with an intact unrelated reference and from 0.0507 to 0.1869 nats
with a shuffled reference.
Without a reference, KL reaches 0.2064 nats
(Table~\ref{tab:dist}), with the same qualitative pattern under
CrossShuffle.

Teacher--student divergence therefore survives semantic disruption, but
its magnitude alone does not explain performance:
shuffling the reference increases KL while reducing final accuracy.
OPSD is thus governed jointly by \emph{where} supervision is collected
and \emph{what} teacher distribution is provided there.

%% file: tables/main_res.tex
\begin{table}[!t]
\caption{Evaluation on intact problems, using each trained condition's final checkpoint (step 50).
Entries are Avg@12 (\%); Avg. is the equally weighted three-benchmark mean.
$\Delta$Ref and $\Delta$OPSD are differences in the reported means relative
to UnrelatedRef and Standard OPSD, respectively, in percentage points.
UnrelatedRef uses unrelated references; OPSD uses target-associated references.
Bold marks the largest value in each score column within a model size.}
\label{tab:main_results}
\centering
\small
\setlength{\tabcolsep}{8pt}
\renewcommand{\arraystretch}{1.2}

\begin{tabular}{@{}llrrrrrr@{}}
\toprule
Model
& Condition
& AIME24
& AIME25
& HMMT25
& Avg.
& $\Delta$Ref
& $\Delta$OPSD \\
\midrule

\multirow{7}{*}{Qwen3-1.7B}
& Frozen Base
& 50.28
& 35.00
& 21.94
& 35.74
& -3.06
& -3.52 \\

& UnrelatedRef
& 52.78
& 39.17
& 24.44
& 38.80
& 0.00
& -0.46 \\

& OPSD
& 53.06
& 38.89
& 25.83
& 39.26
& +0.46
& 0.00 \\

& \textbf{Shuffle-1 (Ours)}
& \textbf{57.22}
& \textbf{45.83}
& 25.28
& \textbf{42.78}
& \textbf{+3.98}
& \textbf{+3.52} \\

& CrossShuffle
& 53.33
& 44.72
& 25.28
& 41.11
& +2.31
& +1.85 \\

& AllShuffle-1
& 51.94
& 42.50
& \textbf{26.67}
& 40.37
& +1.57
& +1.11 \\

& NoTarget
& 56.67
& 43.33
& 25.28
& 41.76
& +2.96
& +2.50 \\

\midrule

\multirow{7}{*}{Qwen3-4B}
& Frozen Base
& 72.78
& 63.89
& 43.33
& 60.00
& -2.50
& -1.95 \\

& UnrelatedRef
& 75.00
& 68.61
& 43.89
& 62.50
& 0.00
& +0.55 \\

& OPSD
& 73.06
& 68.06
& 44.72
& 61.95
& -0.55
& 0.00 \\

& \textbf{Shuffle-1 (Ours)}
& 75.28
& 70.00
& 43.89
& \textbf{63.06}
& \textbf{+0.56}
& \textbf{+1.11} \\

& CrossShuffle
& \textbf{77.22}
& \textbf{70.28}
& 40.83
& 62.78
& +0.28
& +0.83 \\

& AllShuffle-1
& 74.72
& \textbf{70.28}
& 42.22
& 62.41
& -0.09
& +0.46 \\

& NoTarget
& 74.72
& 65.56
& \textbf{45.56}
& 61.95
& -0.55
& 0.00 \\

\midrule

\multirow{7}{*}{Qwen3-8B}
& Frozen Base
& 73.61
& 64.17
& 44.17
& 60.65
& -2.68
& -2.41 \\

& UnrelatedRef
& 76.11
& 68.89
& 45.00
& 63.33
& 0.00
& +0.27 \\

& OPSD
& 75.56
& 70.28
& 43.33
& 63.06
& -0.27
& 0.00 \\

& \textbf{Shuffle-1 (Ours)}
& \textbf{77.78}
& 69.72
& 45.83
& \textbf{64.44}
& \textbf{+1.11}
& \textbf{+1.38} \\

& CrossShuffle
& 74.17
& 69.17
& \textbf{46.94}
& 63.43
& +0.10
& +0.37 \\

& AllShuffle-1
& 76.39
& \textbf{70.56}
& 41.67
& 62.87
& -0.46
& -0.19 \\

& NoTarget
& 75.56
& 65.83
& 45.28
& 62.22
& -1.11
& -0.84 \\

\bottomrule
\end{tabular}
\end{table}

%% file: tables/controls.tex
\begin{table}[htbp]
\caption{Reference and teacher-mode controls on Qwen3-1.7B. All students use non-thinking mode without a problem-specific target. On/Off denotes teacher thinking mode. Each control changes one factor relative to NoTarget. Means are three-benchmark Avg@12 (\%); $\Delta$NoTarget is in percentage points.}
\label{tab:controls}
\centering
\small
\setlength{\tabcolsep}{5.5pt}
\renewcommand{\arraystretch}{1.1}
\begin{tabular}{@{}lccrr@{}}
\toprule
Condition & Reference? & Teacher mode & Mean & $\Delta$NoTarget\\
\midrule
NoTarget & Yes & On & 41.76 & +0.00 \\
NoTarget-NoRef & No & On & 37.78 & -3.98 \\
NoTarget-SameMode & Yes & Off & 38.98 & -2.78 \\
\bottomrule
\end{tabular}
\end{table}

%% file: tables/distribution_core.tex
\begin{table}[htbp]
\caption{Teacher--student divergence on matched Shuffle-1 prefixes. Entries are uncapped $D_{\mathrm{KL}}(p^T\|p^S)$ in nats for the frozen Qwen3-4B base student, excluding the first generated position. $\Delta$KL is thinking minus non-thinking KL, not a divergence between the two teachers.}
\label{tab:dist}
\centering
\small
\setlength{\tabcolsep}{5.5pt}
\renewcommand{\arraystretch}{1.1}
\begin{tabular}{@{}lrrr@{}}
\toprule
Teacher-only reference
& Non-thinking
& Thinking
& $\Delta$KL \\
\midrule
None & 0.0000 & 0.2064 & +0.2064 \\
Intact unrelated & 0.0419 & 0.1605 & +0.1186 \\
Word-shuffled & 0.0507 & 0.1869 & +0.1362 \\
\bottomrule
\end{tabular}
\end{table}

%% file: sections/related_work.tex
\section{Related Work}
\label{sec:related}
\paragraph{Distillation on student trajectories.}
Knowledge distillation transfers predictive distributions \citep{R30}, and generalized distillation connects teacher supervision to privileged information \citep{R18}. Same-architecture students can also benefit from distillation \citep{R32}. Sequence-level distillation uses teacher-generated outputs \citep{seqkd}, whereas GKD and MiniLLM supervise student-generated trajectories \citep{R17,R29}. DistiLLM variants examine response sources and loss design \citep{distillm,distillm2}. The importance of learner-visited states also arises in interactive imitation learning \citep{R31}. We study the shared input that shapes those states and their supervision.

\paragraph{Context-conditioned self-teaching.}
Context distillation transfers context-induced behavior into model parameters \citep{R25,R26}. Related methods learn from privileged solutions \citep{R01}, demonstrations \citep{sdft}, textual feedback \citep{sdpo}, or successful generated rationales \citep{star}. Recent OPSD studies examine reference benefits \citep{R03,R04}, problem-space guidance \citep{R11}, trajectory-conditioned teachers \citep{R05}, and pseudo-solutions \citep{R10}. Work on thinking degradation motivates changes to token-level supervision \citep{R09,R27,R06}. Our design complements these approaches by changing the input shared by student and teacher while retaining a fixed reference assignment.

\paragraph{Text perturbations and prompt structure.}
Word swaps in augmentation \citep{eda}, denoising from corrupted text \citep{bart}, shuffled-text pretraining \citep{word_order_mlm}, and word-order robustness tests \citep{unnatural_nli} provide precedents for disrupting text structure. Demonstration content and order also affect in-context prediction \citep{R20,prompt_order}. Shuffled OPSD uses transformed text to collect and supervise student trajectories, without reconstructing the intact target or assigning it the original answer as a label. The outcome of interest is accuracy after training, evaluated on intact problems.

%% file: sections/conclusion.tex
\section{Conclusion}
\label{sec:discussion_conclusion}

Our results show that semantic coherence is not uniformly beneficial for OPSD. Shuffled OPSD improves reasoning by disrupting the shared problem while preserving a structured teacher-only reference, outperforming the compared conditions across three model scales without changing the distillation objective or inference procedure. The key finding is a structural asymmetry: shuffling the shared problem helps, whereas shuffling the teacher reference hurts, motivating \emph{selective semantic disruption} as a design principle for OPSD---perturb the input that shapes student trajectories while preserving the structure that supports teacher guidance. More broadly, distillation inputs should be designed not only for semantic fidelity, but for the learning dynamics they induce, through both the trajectories they elicit and the supervision they enable.

%% file: sections/ai_use.tex
\section*{AI Use Statement}
The authors defined the research questions, interventions, experimental design, and interpretation of results. Generative AI tools assisted with writing, language editing, and debugging existing code and experimental workflows. The authors are responsible for verifying the measurements, analysis, and final manuscript.

\section*{Reproducibility Statement}
The main paper and appendix specify the data transformations, reference pairing, training configuration, evaluation protocol, and diagnostic aggregation. The complete codebase and model checkpoints will be released upon acceptance.

%% file: sections/appendix.tex
\section{Implementation and Evaluation Details}
\label{app:implementation}

This appendix provides the implementation details shared across the
Qwen3-1.7B, Qwen3-4B, and Qwen3-8B experiments.
Unless explicitly varied by an intervention, all conditions use the
same training data, optimization settings, rollout procedure,
distillation objective, and evaluation protocol.

\subsection{Training Configuration}
\label{app:optimization_details}

\paragraph{Data.}
We use the full OpenThoughts mathematical-reasoning training pool used
by the original OPSD study, containing 29,434 problem--solution pairs.
No fixed 800-example subset is constructed.
All conditions draw training examples from the same seeded ordering of
this pool.

\paragraph{Models and optimization.}
Students initialize from Qwen3-1.7B, Qwen3-4B, or Qwen3-8B with fresh LoRA
adapters.
The backbone remains frozen and only the student adapter is optimized.
The teacher is a frozen copy of the initial backbone and remains fixed
throughout training.
Student rollouts use non-thinking mode, while teacher scoring uses
thinking mode unless explicitly changed by an intervention.

Table~\ref{tab:training_details} summarizes the shared training
configuration.

\begin{table}[H]
\centering
\caption{Shared training configuration for all main experiments.}
\label{tab:training_details}
\small
\renewcommand{\arraystretch}{1.13}
\begin{tabularx}{\linewidth}{@{}l X@{}}
\toprule
Setting & Value \\
\midrule
Backbones
& Qwen3-1.7B; Qwen3-4B; Qwen3-8B \\

Training data
& Full OpenThoughts mathematical-reasoning training pool (29,434 records) \\

Training seed
& 42 \\

Optimizer steps
& 50 \\

Global / generation batch size
& 16 / 16 \\

Optimizer
& AdamW; $\beta_1=0.9$, $\beta_2=0.999$, $\epsilon=10^{-8}$ \\

Learning rate
& $5\times10^{-6}$, constant schedule \\

Warmup / weight decay
& 0 / 0 \\

Gradient clipping
& Global norm 0.1 \\

LoRA
& $r=64$, $\alpha=128$, dropout $0$ \\

LoRA modules
& \texttt{q\_proj}, \texttt{k\_proj}, \texttt{v\_proj},
  \texttt{o\_proj}, \texttt{gate\_proj},
  \texttt{up\_proj}, \texttt{down\_proj} \\

Student / teacher mode
& Non-thinking / thinking \\

Rollout sampling
& Temperature $1.1$, top-$p=0.95$, top-$k=20$ \\

Maximum prompt / response length
& 4,096 / 1,024 tokens \\

Distillation temperature
& 1.1 \\

Pointwise KL cap
& 0.05 \\

Checkpoint interval
& Every 10 optimizer steps \\
\bottomrule
\end{tabularx}
\end{table}

\paragraph{On-policy rollouts.}
At each optimizer step, responses are sampled from the current student
under the condition-specific input.
The same generated continuation is then replayed under the student and
teacher prompts to obtain next-token distributions on identical
prefixes.
Prompt and padding tokens are excluded from the loss.
An EOS token is included when generated; responses reaching the maximum
length are scored without appending an artificial EOS.

\paragraph{Distillation objective.}
Let $p_{it}(v)$ and $q_{it}(v)$ denote the temperature-scaled student
and frozen-teacher probabilities for vocabulary entry $v$ at response
position $t$ of example $i$.
For batch size $B$, the implemented objective is
\begin{equation}
\mathcal{L}_{c}
=
\frac{1}{B}
\sum_{i=1}^{B}
\frac{1}{T_i}
\sum_{t=1}^{T_i}
\sum_{v\in\mathcal{V}}
\min\left\{
q_{it}(v)\log\frac{q_{it}(v)}{p_{it}(v)},
c
\right\},
\qquad c=0.05.
\label{eq:implementation_loss}
\end{equation}

The cap is applied independently to each signed vocabulary contribution
before summation; negative contributions are retained.
The loss is averaged over supervised positions within each response and
then equally across responses.
Teacher probabilities, sampled continuations, and sampling probabilities
are treated as constants during differentiation.
No additional temperature-squared multiplier is applied.

\subsection{Semantic Interventions}
\label{app:intervention_details}

Table~\ref{tab:intervention_details} summarizes the intervention
conditions.
The shared problem is always identical in the student and teacher
prompts; privileged references appear only in the teacher prompt.

\begin{table}[t]
\centering
\caption{
Semantic interventions used in the experiments.
Only the listed conditioning component differs from the matched control.
}
\label{tab:intervention_details}
\small
\setlength{\tabcolsep}{3.8pt}
\renewcommand{\arraystretch}{1.18}
\begin{tabularx}{\linewidth}{@{}l X X l@{}}
\toprule
Condition
& Shared problem
& Teacher-only reference
& Teacher mode \\
\midrule

Standard OPSD
& Intact
& Associated solution
& Thinking \\

UnrelatedRef
& Intact
& Unrelated worked example
& Thinking \\

Shuffle-1
& Word shuffled
& Same as UnrelatedRef
& Thinking \\

Shuffle-4
& Four-word blocks shuffled
& Same as UnrelatedRef
& Thinking \\

Shuffle-8
& Eight-word blocks shuffled
& Same as UnrelatedRef
& Thinking \\

CrossShuffle
& Words redistributed across problems
& Same as UnrelatedRef
& Thinking \\

NoTarget
& Problem-specific content removed
& Same as UnrelatedRef
& Thinking \\

AllShuffle-1
& Word shuffled
& Unrelated example word shuffled
& Thinking \\

NoTarget-NoRef
& Same as NoTarget
& None
& Thinking \\

NoTarget-SameMode
& Same as NoTarget
& Same as UnrelatedRef
& Non-thinking \\
\bottomrule
\end{tabularx}
\end{table}

\paragraph{Unrelated references.}
Let the training pool contain $N=29{,}434$ records.
For training record $i$, UnrelatedRef uses record
\begin{equation}
(i+111)\bmod N
\end{equation}
as the teacher-only worked example.
The nonzero cyclic shift prevents self-pairing and is fixed across all
matched target-side interventions.

\paragraph{Within-problem shuffling.}
Shuffle-$k$ partitions the problem into consecutive blocks of $k$
whitespace-delimited words and permutes the block order.
Shuffle-1 therefore permutes individual words, while larger $k$
preserves progressively longer local spans.
The transformation preserves the word multiset but need not preserve
tokenizer-token counts.

\paragraph{Cross-problem shuffling.}
CrossShuffle pools words across the training problems, applies a fixed
global permutation, and redistributes them according to the original
per-example word counts.
It preserves corpus-level lexical counts and example lengths while
breaking problem membership.

\paragraph{Reference shuffling and target removal.}
AllShuffle-1 applies word-level shuffling separately to the shared
problem, the unrelated reference problem, and its solution.
These fields are never pooled together.
NoTarget removes the problem-specific statement and its wrapper while
retaining the general instruction.
NoTarget-NoRef additionally removes the teacher reference.
Generated student continuations are never shuffled.

\subsection{Evaluation Protocol}
\label{app:evaluation_details}

We evaluate AIME 2024, AIME 2025, and HMMT February 2025, with
30 problems per benchmark.
Evaluation always uses the original intact problem statements and no
teacher-only reference.
For every trained condition, we evaluate and report the final checkpoint
after 50 optimizer steps under the same protocol.

\begin{table}[t]
\centering
\caption{Benchmark evaluation protocol.}
\label{tab:evaluation_details}
\small
\renewcommand{\arraystretch}{1.15}
\begin{tabularx}{\linewidth}{@{}l X@{}}
\toprule
Setting & Value \\
\midrule

Benchmarks
& AIME 2024; AIME 2025; HMMT February 2025 \\

Problems per benchmark
& 30 \\

Input
& Original intact problem \\

Teacher-only reference
& None \\

Thinking mode
& Enabled \\

Temperature
& 1.0 \\

Top-$p$ / top-$k$
& 0.95 / disabled \\

Maximum new tokens
& 38,912 \\

Responses per problem
& 12 \\

Responses per checkpoint
& 1,080 across all three benchmarks \\

Reported checkpoint
& Final saved checkpoint after 50 optimizer steps for every condition \\

Question ordering
& Fixed shuffle with seed 2026 \\

Generation seed
& $s_i=2026+i$ for shuffled position $i$ \\

Primary metric
& Avg@12 \\

Benchmark aggregation
& Equal weight across the three benchmarks \\
\bottomrule
\end{tabularx}
\end{table}

\paragraph{Final-checkpoint evaluation.}
Checkpoints are saved every 10 optimizer steps. All trained conditions
are evaluated at the final checkpoint after 50 optimizer steps.
Intermediate checkpoints and benchmark scores are not used for
checkpoint selection; the frozen-base rows use the untrained backbone.

\paragraph{Evaluation metric.}
Let $R_{i,j}\in\{0,1\}$ indicate whether response $j$ to problem $i$ is
correct.
For $N$ problems,
\begin{equation}
\mathrm{Avg@12}
=
\frac{1}{12N}
\sum_{i=1}^{N}
\sum_{j=1}^{12}
R_{i,j}.
\end{equation}
The reported overall mean assigns equal weight to the three benchmarks.

\paragraph{Seed schedule.}
Within each benchmark, problems are shuffled once using seed 2026.
The problem at zero-based shuffled position $i$ uses generation seed
$s_i=2026+i$.
All compared conditions share the same problem ordering and seed
schedule.

\subsection{Statistical Estimation}
\label{app:statistics_details}

For paired comparisons with saved response-level outputs, conditions
are aligned by problem ID and evaluated using 10,000 problem-cluster
bootstrap resamples.
All 12 responses from a problem remain in the same cluster, and paired
conditions use identical resampled problem indices.
For three-benchmark summaries, resampling is performed within each
benchmark before computing the equal-weight mean.

The reported 95\% intervals quantify uncertainty across evaluation
problems conditional on the final checkpoints and generated
responses.
Main experiments use a single training seed, so these intervals do not
estimate training-seed variability.
Comparisons without matched response-level outputs are reported as point
differences.

\section{Sensitivity to Shuffle Granularity}
\label{app:sensitivity}

We evaluate how the granularity of semantic disruption shapes the
effectiveness of Shuffled OPSD.
Shuffle-$k$ partitions each shared problem into consecutive blocks of
$k$ whitespace-delimited words and permutes these blocks, so larger
$k$ preserves progressively more local word order.
AllShuffle-$k$ applies the same transformation to both the shared
problem and the teacher-only reference.

Table~\ref{tab:shuffle_sensitivity} shows a clear pattern.
For Qwen3-1.7B, Shuffle-4 and Shuffle-8 achieve three-benchmark means
of 41.02\% and 40.65\%, respectively, both below the 42.78\%
obtained by Shuffle-1 in Table~\ref{tab:main_results}.
For Qwen3-4B, Shuffle-4 and Shuffle-8 achieve 62.22\% and 62.87\%,
respectively, compared with 63.06\% for Shuffle-1.
Across both model sizes, the strongest target-side intervention is
therefore the most aggressive one: disrupting word order at the
individual-word level.
Preserving additional local structure does not improve performance and,
at 1.7B, weakens it substantially.

The AllShuffle variants reveal the complementary effect on the
teacher-only reference.
At 1.7B, increasing the block size from 4 to 8 raises the mean from
38.80\% to 40.37\%, whereas at 4B it lowers the mean from 62.13\% to
61.48\%.
More importantly, every AllShuffle variant underperforms its matched
target-only Shuffle counterpart.
The benefit of semantic disruption is therefore strongly
location-dependent: aggressive disruption is most effective on the
shared problem, while preserving structure in the teacher-only reference
consistently produces stronger students.

Taken together, these results strengthen the central design principle of
Shuffled OPSD.
The useful signal does not come from preserving progressively more local
semantic structure in the shared problem.
Instead, performance is maximized by strongly disrupting the shared
problem while retaining a structured teacher context.

\input{tables/sensitive}

%% file: tables/sensitive.tex
\begin{table}[htbp]
\caption{Sensitivity to shuffle granularity. Entries are Avg\@12 (\%); Avg. is the equally weighted three-benchmark mean.}
\label{tab:shuffle_sensitivity}
\centering
\small
\setlength{\tabcolsep}{5.5pt}
\renewcommand{\arraystretch}{1.1}
\begin{tabular}{@{}llrrrr@{}}
\toprule
Model
& Condition
& AIME24
& AIME25
& HMMT25
& Avg. \\
\midrule

\multirow{4}{*}{Qwen3-1.7B}
& Shuffle-4
& 54.17
& 41.39
& 27.50
& 41.02 \\

& Shuffle-8
& 56.67
& 41.94
& 23.33
& 40.65 \\

& AllShuffle-4
& 52.78
& 38.33
& 25.28
& 38.80 \\

& AllShuffle-8
& 55.83
& 41.39
& 23.89
& 40.37 \\

\midrule

\multirow{4}{*}{Qwen3-4B}
& Shuffle-4
& 76.67
& 66.94
& 43.06
& 62.22 \\

& Shuffle-8
& 75.28
& 68.61
& 44.72
& 62.87 \\

& AllShuffle-4
& 73.89
& 69.17
& 43.33
& 62.13 \\

& AllShuffle-8
& 74.44
& 68.06
& 41.94
& 61.48 \\

\bottomrule
\end{tabular}
\end{table}